\documentclass[runningheads]{llncs}
\usepackage{graphicx}
\usepackage{amsmath}
\usepackage{booktabs}
\begin{document}
\title{Deep Learning for Bias Detection:\\
From Inception to Deployment}
%
%
\author{Md Abul Bashar\inst{1}\orcidID{0000-0003-1004-4085}\and
Richi Nayak\inst{1}\and
Anjor Kothare\inst{2}\and
Vishal Sharma\inst{2}\and
Kesavan Kandadai\inst{2}
}
%
%
\institute{Queensland University of Technology, Brisbane, Australia\\
\email{\{m1.bashar, r.nayak\}@qut.edu.au} \and
ishield.ai\\
\email{anjor.kothare@ishield.ai, vishal.sharma@tothenew.com, kesavan@pallavatech.com, kesavan@ishield.ai}
}
\maketitle              
\begin{abstract}
To create a more inclusive workplace, enterprises are actively investing in identifying and eliminating unconscious bias (e.g., gender, race, age, disability, elitism and religion) across their various functions. We propose a deep learning model with a transfer learning based language model to learn from manually tagged documents for automatically identifying bias in enterprise content. We first pretrain a deep learning-based language-model using Wikipedia, then fine tune the model with a large unlabelled data set related with various types of enterprise content. Finally, a linear layer followed by softmax layer is added at the end of the language model and the model is trained on a labelled bias dataset consisting of enterprise content. The trained model is thoroughly evaluated on independent datasets to ensure a general application. We present the proposed method and its deployment detail in a real-world application.

\keywords{Deep Learning \and Bias Detection \and Transfer Learning \and Text Data \and Small Dataset.}
\end{abstract}

\section{Introduction}
\vspace{-2mm}
A rigorous study by McKinsey \cite{Dixon-Fyle2020DiversityMatters} found that more diverse and inclusive organisations outperform those that are not. The concept of unconscious bias has become increasingly pervasive, with many organisations training their employees on the concept of Diversity and Inclusion (D\&I). Unfortunately, though an increased awareness of unconscious bias can have benefits, it is not a systemic and consistent solution \cite{Herbert2021IsWorthwhile}.
Enterprises are looking for technologies that can create consistent and scalable practices to identify or mitigate biases across organisations, often in real-time.

While there have been a few analytics products to measure employee demographics, pay parity and customised training \cite{Mitchell2020DiversitySelection}, there is lack of solutions targeted toward tackling unconscious bias in content that is created within an enterprise. For every enterprise, almost always the first touch point for their users, vendors, employees or business partners is the content published across their digital channels. This content may be job descriptions, website content, marketing messages, blogs, reports and social media content. Internally within the enterprise, content is also created in the form of product documentation, user guides, customer care, messaging and collaboration platforms. 
Unconscious bias in content manifests itself in various forms in different types of content. For example: (a) in job descriptions this could be through use of pronouns or masculine coded words, (b) in enterprise messaging platforms this could be through toxic messages and bullying, (c) in product descriptions this could be through stereotyping a type of a user, (d) in e-commerce product details this could be through references to body shapes, or (e) in a stock market report, this could be through inherent assumption of the gender of a potential investor.

In this paper, we aim to propose a Deep Learning (DL) model to detect unconscious bias in content and how it can be applied to integrate and work with enterprise applications. Lexical based systems and traditional machine learning systems do not solve the problem due to complexity and intricate relationships inherent in the textual narratives. With advancements in DL in Natural Language Processing (NLP), it is feasible to build a DL based system. However, such a system requires a large set of labelled data for building the model. With the manual efforts involved in labelling the data, it is difficult to create a large dataset. There do not exist similar data that can be used in labelling. In this paper we propose a novel method based on transfer learning to deal with the small set of labelled data and detect unconscious bias in content.

This research makes the following novel contributions.
(1) It proposes a comprehensive bias detection model that can detect four types of bias (\emph{Race}, \emph{Gender}, \emph{Age}, \emph{Not Appropriate}) in text data.
(2) It uses progressive transfer learning that allows to train a smaller model (i.e. less number of hyper parameters) with a small labelled dataset for better accuracy. 
(3). It presents the detail of deploying the model in a real-world application\footnote{https://ishield.ai/}.

\vspace{-6mm}
\section{Literature review}
\vspace{-4mm}

When the unconscious biases are not tackled, they cause serious harm to the business including financially, socially and culturally \cite{2020TheZ,Zalis2019InclusiveResearch,Bailinson2020UnderstandingWorkplace,Dixon-Fyle2020DiversityMatters,JohnsonIfHired,Agovino2019ToxicBillions}. 
However, very few to no technology solutions exist to tackle unconscious bias in the content that is created and published across the enterprise for both their internal and external audiences \cite{GarrStaciaShermanandJackson2019DiversityMarket}.

Deep learning models have become quite successful in natural language processing, e.g., content generation, language translation, Question and Answering systems, text classification \cite{Bashar2019MisogynisticDatasets,Bashar2019QutNocturnalHASOC19:Languageb,Bashar2020RegularisingSet,Bashar2021ProgressivePosts,Abul2021ActiveTask} and clustering. In spite of this success of DL in NLP, there has been very limited works on building a generic bias detection method. The work in current literature can be grouped into two categories: (1) text representation learning; and (2) reducing subjective bias in text.

Researchers\footnote{https://gender-decoder.katmatfield.com/} \cite{Manzini2019BlackEmbeddings,Bordia2019IdentifyingModels} proposed to debias semantic representation of words by removing bias component from word embedding. The goal is to make semantic representations fair across attributes like gender and age. Autoencoder was used to generate a balanced gender-oriented word distribution to remove gender bias from word embeddings\cite{Kaneko2019Gender-preservingEmbeddings}. Counterfactual data has been augmented to alter the training distribution to balance gender-based statistics \cite{Maudslay2019ItsSubstitution,Zmigrod2019CounterfactualMorphology,Dinan2020QueensGeneration}. Research in \cite{WangBalancedRepresentations} used adversarial training to squeeze directions of bias in the hidden states of image representation. 

Research in \cite{Pryzant2020AutomaticallyText} proposed a model for automatically suggesting edits for subjective-bias words following the Wikipedia’s neutral point of view (NPOV) policy for defining subjective bias. However, the model is limited to single-word edits, i.e. it can handle simplest instances of bias only. Research in \cite{Hube2018DetectingWikipedia,Recasens2013LinguisticLanguage} used lexicon of bias words for detecting language bias in sentences of Wikipedia articles. A fine-tuned BERT model \cite{devlin2018bert} is used for detecting gender bias only \cite {DinanMulti-DimensionalClassification}. 

An issue faced by bias detection methods in multi-class setting is failing to select minority class examples in imbalanced data distribution \cite{attenberg2011inactive}. Guided learning, based on crowd-sourcing to find or generate class-specific training instances, can help to get more balanced class frequencies \cite{patterson2015tropel,attenberg2010label}. However, guided learning is resource consuming and may not present the true distribution generating training examples. Sometimes, heuristic labelling methods such as distant supervision \cite{craven1999constructing} or data programming \cite{ehrenberg2016data} are used for datasets with the imbalanced class distribution. However, these methods are only applicable when a good knowledge base or a pretrained predictor is available \cite{c2018active}. 

In this research, we propose to use progressive transfer learning to address the class imbalance problem. To our best of knowledge, there exist no model that comprehensively detect common biases (e.g. race, gender, age) in text.


\vspace{-6mm}
\section{Language based Deep Learning model for Bias Detection}
\vspace{-4mm}
Bias detection in the text data is a complex problem because usually bias is represented by linguistic cues that are subtle and can be determined only through its context in the text. 
Let $X$ be a text dataset that contains $n$ features and $K$ classes. Let $\mathbf{x} = \langle x_1,\dots x_n \rangle$ be a vector representing an instance in $X$. Let $C_k$ be a set of $K$ classes. The bias detection is a classification task that assigns an instance to a bias class (or category) $C_k$ based on the feature vector $\mathbf{x}$; i.e. $f \in \mathcal{F}: X \rightarrow C_k$, where $f(\mathbf{x}) = \max_{C_k} \, p(C_k|\mathbf{x})$. This ascertains that we need to know $p(C_k|\mathbf{x})$ for a bias detection task. 
The joint probability $p(\mathbf{x}, C_k)$ of $\mathbf{x}$ and $C_k$ can be written as
\begin{equation}
    p(\mathbf{x}, C_k) = p(C_k|\mathbf{x})p(\mathbf{x})
\label{eq:bayes_simp_rearr}
\end{equation}
where $p(\mathbf{x})$ is the prior probability distribution. The prior probability $p(\mathbf{x})$ can be seen as a regulariser for $p(C_k|\mathbf{x})$ that can regularise modelling of the associated uncertainties of $p(\mathbf{x}, C_k)$ \cite{Bashar2020RegularisingSet}. As $p(\mathbf{x})$ does not depend on $C_k$, this means that $p(\mathbf{x})$ can be learned independent of the class level $C_k$. That is, $p(\mathbf{x})$ can be learned from unlabelled data. 

Prior research \cite{Bashar2020RegularisingSet,Bashar2021ProgressivePosts} showed that the estimation of $p(\mathbf{x}, C_k)$ can be improved when $p(\mathbf{x})$ is learned from a sequence of unlabelled datasets, especially when the labelled dataset is small. In this study, we propose to implement a bias classification model utilising this technique of improving the prediction accuracy through unlabelled data.

A discriminative model such as LSTM learns to classify an instance $\mathbf{x}$ into class $C_k$ by learning the conditional probability distribution as $p(C_k|\mathbf{x}, \theta) \approx p(C_k|\mathbf{x})$ where $\theta$ is the list of model parameters.
However, accurately approximating $p(C_k|\mathbf{x})$ requires a large number of labelled instances. If only a small set of labelled data is available, the learned $p(C_k|\mathbf{x}, \theta)$ might not be a good approximation of population distribution because $\theta$ may over-fit the small training set. 
Alternately, $p(\mathbf{x})$ can be learned from one or more large unlabelled datasets and conditioned on $\mathbf{x}$ to learn $p(C_k|\mathbf{x}, \theta)$, leading to $p(C_k|\mathbf{x}, \theta)p(\mathbf{x})$ $\approx$ $p(\mathbf{x}, C_k)$. The term $p(C_k|\mathbf{x}, \theta)p(\mathbf{x})$ can be seen as equivalent to combining the regularisation into the discriminative model. This regularised model would act similar to a generative model. 
	
Unlike common transfer learning where $p(\mathbf{x})$ is learned once from a large unlabelled dataset, we propose to progressively learn $p(\mathbf{x})$ from a sequence of unlabelled datasets. This allows us to use a relatively small set of parameters in our model. Then we use $p(\mathbf{x})$ to learn $p(C_k|\mathbf{x}, \theta)p(\mathbf{x})$ $\approx$ $p(\mathbf{x}, C_k)$ with a small training dataset. Next, we  present the estimation of $p(\mathbf{x})$ as a neural network language model (NNLM) using unsupervised learning.

\vspace{-4mm}
\subsection{Neural Network Language Model}
\vspace{-2mm}
Probability $p(\mathbf{x})$ can be estimated using the assumption of Language model where features are considered conditionally dependent \cite{Bashar2021ProgressivePosts,Bashar2020RegularisingSet}. This is to support natural language processing where in a sentence, the sequencing of words depends on each other. Based on this, the joint probability $p(\mathbf{x},C_k)$ in Equation \ref{eq:bayes_simp_rearr} can be rewritten as follows, using the chain rule:
\begin{equation}
	\begin{aligned}
	p(\mathbf{x},C_k)
	&= p(C_k|\mathbf{x})p(\mathbf{x})\\
	&= p(C_k|\mathbf{x}) p(x_1, \dots, x_n)\\
	&= p(C_k|\mathbf{x}) \prod_{i=1}^n p(x_i|x_1 \dots x_{i-1})
	\end{aligned}
	\label{eq:chain_join}
\end{equation}
The part $\prod_{i=1}^n p(x_i|x_1 \dots x_{i-1})$ in Equation \ref{eq:chain_join} can be considered as a language model because $p(x_i|x_1 \dots x_{i-1})$ seeks to predict the probability of observing the $i$th feature $x_i$, given the previous $(i-1)$ features $(x_1 \dots x_{i-1})$. A Recurrent Neural Network (RNN) or its variants such as Long Short-Term Memory (LSTM) can be used to model $\prod_{i=1}^n p(x_i|x_1 \dots x_{i-1})$ as they work in a similar way to capture the order of features and their non-linear and hierarchical interactions \cite{mikolov2010recurrent,jozefowicz2016exploring,Bashar2020RegularisingSet}. Similar to traditional language models, a RNN/LSTM based NNLM can approximate joint probabilities over the feature sequences as follows, where $\omega$ is the list of model parameters.
	\begin{equation}
	\begin{aligned} 
	p(\mathbf{x}) &= p(x_1, \dots, x_n)\\
	&\approx p(x_1, \dots, x_n, \mathbf{\omega})\\
	&\approx \prod_{i=1}^n p(x_i|x_1 \dots x_{i-1}, \mathbf{\omega})
	\end{aligned}
	\label{eq:lm_rnn}
	\end{equation}

Given a sequence of features, a RNN recurrently processes each feature and uses multiple hidden layers to capture the order of features and their non-linear and hierarchical interactions. The hidden state is used to derive a vector of probabilities representing the network's guess of the subsequent feature in the sequence \cite{Bashar2020RegularisingSet,Bashar2021ProgressivePosts}. The network aims to minimize the loss calculated based on the vector of probabilities and the actual next feature. In simple words, the context of all previous features in the sequence is encoded within the parameters $\mathbf{\omega}$ of the network and the probability of getting the next word is distributed over the vocabulary using a Softmax function \cite{jozefowicz2016exploring}.

Estimating $p(\mathbf{x})$ using a LSTM-based LM (i.e. NNLM) model on a huge dataset that covers multitude of domains is useful for transfer learning, but it can be very expensive in terms of required computation and memory \cite{bradbury2016quasi}. Additionally, it can learn irrelevant and misleading relationships in data due to interactions between different domains in a single corpus \cite{Bashar2020RegularisingSet}. Therefore, as suggested in \cite{Bashar2021ProgressivePosts}, we use an alternative approach based on progressive transfer learning to incorporate knowledge gained from a sequence of datasets in a LSTM-based LM \cite{Bashar2020RegularisingSet}. 

Let there be $m$ number of  corpora from which the knowledge is gained. A LSTM model built on corpus $D_i$ to learn $p(\mathbf{x}|D_i)p(D_i)$ will have its parameters $\omega_i$. It can be expressed as follows.
	\begin{equation}
	\begin{aligned}
	\prod_{i=1}^m p(\mathbf{x}|D_i)p(D_i)
	&\approx \prod_{i=1}^m p(\mathbf{x}|D_i, \omega_i)p(D_i, \omega_i)\\
	&= \prod_{i=1}^m p(\mathbf{x}|D_i, \omega_i)p(\omega_i|D_i)p(D_i)
	\end{aligned}
	\end{equation}
	If the same LM model is sequentially built from the given $m$ datasets, parameters $\omega_i$ learned on $i^{th}$ dataset will only depend on the parameters $\omega_{i-1}$ learned on the $(i-1)^{th}$ dataset, applying the Markov assumption.
	\begin{equation}
	\begin{aligned}
   \prod_{i=1}^m p(\mathbf{x}|D_i, \omega_i)p(\omega_i|D_i)p(D_i)
	& \approx \prod_{i=1}^m p(\mathbf{x}|D_i, \omega_i)p(\omega_i|D_i, \omega_{i-1})p(D_i)
	\end{aligned}
	\end{equation}
	Here $\omega_0$ is the initial weight that might be assigned randomly. Assuming the same probability (or uncertainty) for each dataset, transfer learning can be expressed as follows. 
	\begin{equation}
	\begin{aligned}
	p(\mathbf{x}, D_1, \dots, D_n)
	& \approx \prod_{i=1}^m p(\mathbf{x}|D_i, \omega_i)p(\omega_i|D_i, \omega_{i-1})p(D_i)\\
	&= \prod_{i=1}^m p(\mathbf{x}|D_i, \omega_i)p(\omega_i|D_i, \omega_{i-1})\\
	& \propto \sum_{i=1}^m ln\left(p(\mathbf{x}|D_i, \omega_i)p(\omega_i|D_i, \omega_{i-1})\right)
	\end{aligned}
	\label{eq:mlt_datasets}
	\end{equation}
	Followings can be inferred from Equation \ref{eq:mlt_datasets}. (1) Each dataset $D_i$ relevant to the application domain of LM can reduce uncertainty \cite{Bashar2020RegularisingSet,bashar2018cnn}. (2) Pr-training of LSTM-LM should be done by the order of the dataset of general population distribution to the dataset of specific population distribution because the parameter vector $\omega_i$ depends on $\omega_{i-1}$ \cite{Bashar2021ProgressivePosts,Bashar2020RegularisingSet}. For example, we can approximate the population distribution of Queensland (i.e., specific) from that of Australia (i.e., general) but the opposite is not true.

\vspace{-4mm}
\subsection{Regularising Classifier with Language Model}
\vspace{-2mm}
For the downstream task of classification, a LSTM is trained to learn $p(C_k, \mathbf{x})$ $\approx$ $p(C_k|\mathbf{x}, \theta)p(\mathbf{x})$ on a small training dataset. The prior distribution $p(\mathbf{x})$ can be learned by sequentially pretraining an LSTM on $m$ unlabelled datasets from general to specific domain. Using Equation \ref{eq:mlt_datasets}, we can write the regularised classifier as follows.
\begin{equation}
    \begin{aligned}
    p(C_k, \mathbf{x}) &\approx p(C_k, \mathbf{x}) p(\mathbf{x}, \omega)\\
        &\approx p(C_k|\mathbf{x}, \theta) \prod_{i=1}^m p(\mathbf{x}|D_i, \omega_i)p(\omega_i|D_i, \omega_{i-1})
    \end{aligned}
\end{equation}

\begin{figure}[htb]
	\centering
	\scriptsize
	\includegraphics[width=0.7\textwidth]{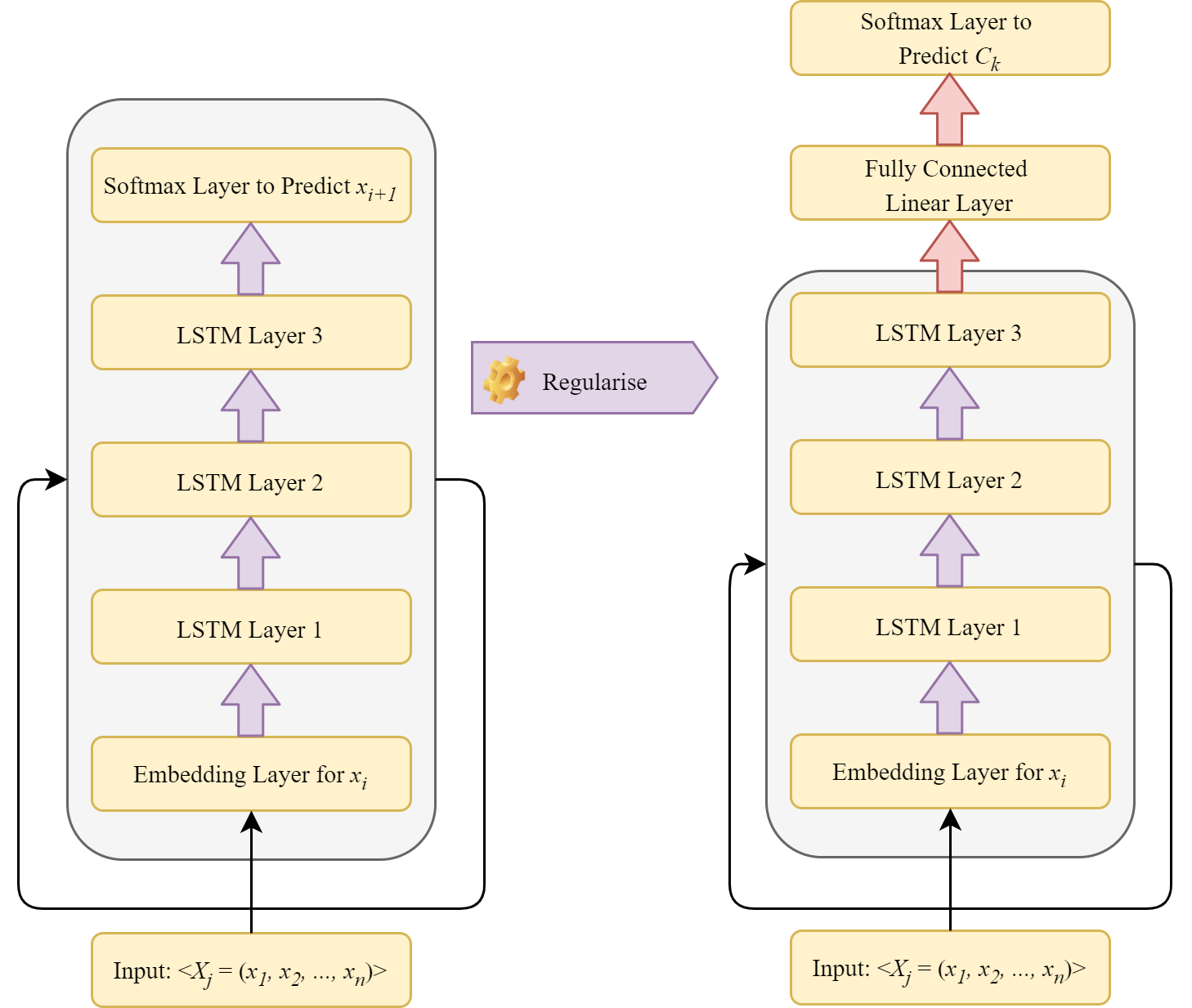}
	\caption{LSTM-L: Process of regularising a classifier using a language model}
	\label{fig:LSTM-L}
	\vspace{-4mm}
\end{figure}

	
Figure \ref{fig:LSTM-L} shows the process of transfer learning through LSTM-based LM to a LSTM classification model. Layers 1 to 3 are stacked LSTM layers. The LSTM-based LM (the left hand side model in Figure \ref{fig:LSTM-L}) is generated using three staked layers along with embedding layer and LM softmax layer. The LM softmax layer is active during pretraining of LM with a sequence of unlabelled datasets and then it is freezed. Once the LM is pretrained, the class softmax layer and linear layer are augmented, as shown by the right hand side model in Figure \ref{fig:LSTM-L}. These two layers along with the pretrained LM active layers are trained with the small labelled dataset to learn the classification task of bias detection. 

The main task of these additional two layers is to learn $p(C_k|\mathbf{x}, \theta)$. The combined network learns $p(C_k|\mathbf{x}, \theta)p(\mathbf{x}, \omega)$. The additional two layers are augmented at the end of NNLM in this model to assure that $\theta$ is learned from the finetuning of $\omega$ with labelled dataset during classification training. 
This process of training a classifier model (e.g. LSTM) generates a classifier regularised by the language model (LM) based transfer learning \cite{Bashar2020RegularisingSet}. We call the trained model as LSTM-LM.

\vspace{-4mm}
\section{Empirical Analysis}
\vspace{-2mm}
Extensive experiments were conducted to evaluate the accuracy of the proposed method for bias detection. We used six standard classification evaluation measures \cite{Bashar2020RegularisingSet}: Accuracy (Ac), Precision (Pr), Recall (Re), F$_1$ Score (F$_1$), Cohen Kappa Score (CKS) and Area Under Curve (AUC).

\vspace{-4mm}
\subsection{Data Collection}
\vspace{-2mm}
\subsubsection{The iShield.ai Dataset}
(Version 1\footnote{In the next versions, this dataset has been considerably modified.}) has a total data count of 57,424 sentences, where 27,131 sentences are biased (47\%) and 30,293 sentences are unbiased (53\%). There are four different kinds of biases: (1) 20,690 (36\%) sentences are GENDER biased, (2) 4,339 (7.56\%) sentences are RACE biased, (3) 1,553 (2.7\%) sentences are ambiguous and (4) 549 (0.96\%) sentences are AGE biased. We used 80-10-10\% ratio between training, validation and testing instances. The dataset comes from two sub-domains namely Job Description (JD) and Non Job Description (NJD) that contributed 25,123 (43.75\%) and 32,301 (56.25\%) sentences respectively.

We used the following four definitions of bias in this research. \textbf{GENDER} -- Any conscious or unconscious attempt to single out a particular gender or people who identify with a particular gender identity. Example: The claims assistant will handles the claims of veterans and their wives. \textbf{RACE} -- Any conscious or unconscious attempt to single out a particular race or it’s people. Example: Own development of brownbag sessions and facilitate publishing reusable content to the organisation. \textbf{AGE} -- Any conscious or unconscious attempt to single out a particular age bracket or it’s people. Example: We are a young organisation looking for young and talented marketers. \textbf{Ambiguous} -- Any part of a sentence which does not clearly convey the intended meaning. Example: We are looking for a smart candidate for this position. 

The basic unit of division used for annotation is a sentence. Any document is first split in sentences. 
Then sentences are passed to annotators in batches, where each batch consists of 150 sentences. Biases are usually observed as a group of four to five words in a sentence. 
A three member panel is set up for a Quality Assurance pass. The panel evaluates the labelled sentences for two checks. 
(1) The labelled sentences are adhering to the outlined definitions of bias. (2) Conflicting instances (e.g. a sentence should have been labelled as biased but is not) are eliminated or placed in the correct category. 

\subsubsection{Sexist Statement in Workplace (SSW) Dataset} \cite{Grosz2020AutomaticWorkplace}
This dataset was collected to check how effectively the classification model works on other datasets besides the bias detection dataset. The SSW dataset has around 1100 labelled instances for sexism statement in workplace. The instances are roughly balanced between sexism (labelled 1) and neutral (labelled 0) cases. Some examples from this dataset is given in Table \ref{tab:sexist_dataset}. We used 80-20\% ratio between training and testing instances. Hyper parameters were set by using cross validation in the 80\% data used for training. 

\begin{table}[htbp]
  \centering
  \scriptsize
  \caption{Instances from SSW Dataset}
    \begin{tabular}{cp{10cm}}
    \toprule
    \multicolumn{1}{l}{Label} & Statement \\
    \midrule
    1     & Women always get more upset than men. \\
    1     & The people at work are childish. it’s run by women and when womendont agree to something, oh man. \\
    1     & I’m going to miss her resting bitch face. \\
    0     & No mountain is high enough for a girl to climb. \\
    0     & It seems the world is not ready for one of the most powerful andinfluential countries to have a woman leader. So sad. \\
    0     & Can you explain why what she described there is wrong? \\
    \bottomrule
    \end{tabular}%
  \label{tab:sexist_dataset}%
  \vspace{-10mm}
\end{table}%

\subsubsection{Model Pretraining Datasets}
We collected a list of pretrained word vectors from \cite{mikolov2018advances}. The list has one million word vectors trained on Wikipedia 2017, UMBC webbase corpus and statmt.org news dataset (16B tokens). We use the following three corpora ($D_1$, $D_2$, and $D_3$) for sequentially pretraining LSTM-LM model and fineturning to target task, i.e. first LSTM-LM is pretrained using $D_1$, then using $D_2$ and finally fintuned to target task using $D_3$. 
	\begin{itemize}
		\item[] $D_1$: The goal of using this corpus is to capture general properties of the English language. We pretrain the LSTM-LM model on Wikitext-103 that contains 28,595 preprocessed Wikipedia articles and 103 million words \cite{merity2016pointer}. After pretraining on $D_1$, we approximate the probability distribution $p(\mathbf{x}|D_1,\mathbf{\omega}_1)$.
		
		\item[] $D_2$: The goal of using this corpus is to bridge the data distribution between the target task domain (i.e., bias detection in job description) and the general domain (i.e. standard language). This is because the target task is likely to come from a different distribution than the general corpus. $D_2$ should be chosen such that it has commonalities with both $D_1$ reflecting a general domain (Wikipedia) and the corpus $D_3$ reflecting a target domain (e.g. labelled data of job description). We use a set of unlabelled JD and NJD data as $D_2$.
		
\end{itemize}




\subsection{Baseline Models}
	We have implemented 10 baseline models to compare the performance of the proposed LSTM-LM.
	
	Models with pretrained word vectors (i.e. word embeddings by word2Vec) include (1) LSTM with pretrained Word vectors (LSTM-W) \cite{hochreiter1997long} and (2) CNN with pretrained Word vectors (CNN-W) \cite{bashar2018cnn}. LSTM-W has 100 units, 50\% dropout, binary cross-entropy loss function, Adam optimiser and sigmoid activation. The hyper parameters of CNN-W is set as in \cite{bashar2018cnn}. We used one million pretrained word vectors each with 300-dimension \cite{mikolov2018advances}. Word vectors are pretrained on Wikipedia 2017, UMBC webbase corpus and statmt.org news dataset. A Continuous Bag-of-Words Word2vec \cite{mikolov2013distributed} model is used in pretraining.
		
	LSTM and CNN without pretrained word vectors (LSTM-P) \cite{hochreiter1997long,bashar2018cnn}. These are traditional LSTM and CNN models that have not been pretrained by any data. Similar to LSTM-W, LSTM-P has 100 units, 50\% dropout, binary cross-entropy loss function, Adam optimiser and sigmoid activation. The hyper parameters of CNN is set as in \cite{bashar2018cnn}.
		
	Feedforward Deep Neural Network (DNN) \cite{glorot2010understanding}. It has five hidden layers, each layer containing eighty units, 50\% dropout applied to the input layer and the first two hidden layers, softmax activation and 0.04 learning rate. For all neural network based models, hyperparameters are manually tuned based on cross-validation. 
		
	Non NN models include Support Vector Machines (SVM) \cite{Hearst1998}, Random Forest (RF) \cite{liaw2002classification}, Decision Tree (DT) \cite{Safavian1991AMethodology}, 
	Gaussian Naive Bayes (GNB) \cite{lewis1998naive}, k-Nearest Neighbours (kNN) \cite{weinberger2009distance} and Ridge Classifier (RC) \cite{hoerl1970ridge}. Hyperparameters of all these models are automatically tuned using ten-fold cross-validation and GridSearch using sklearn library. 
		
	None of the models, except LSTM-LM, LSTM-W and CNN-W, are pretrained or utilised unlabelled dataset.

\subsection{Experimental Results: SSW Dataset}
  \vspace{-2mm}
SSW is a very small dataset. The experimental results on SSW dataset are given in Table \ref{tab:ssw_results}. The proposed language model-based transfer learning model LSTM-LM outperforms all the baseline models. Beside LSTM-LM, other two word vector-based transfer learning-based models, CNN-W and LSTM-W yield the second and third best performance, respectively. We conjecture that (1) transfer learning-based models produce better outcome, and (2) the language model-based transfer learning brings more benefits than the word vector-based transfer learning when the training dataset is small. 

It is interesting to note that CNN-W outperforms CNN as well as LSTM-W outperforms LSTM. CNN-W and CNN (or LSTM-W and LSTM) use the exactly same architecture, except that CNN-W (or LSTM-W) uses pretrained word vectors for transfer learning. This further emphasises the benefit of using transfer learning over the standard models when the training dataset is small. 

Traditional models (i.e. RF, DT, GNB and kNN) do not utilise any transfer learning and solely rely on the labelled training dataset. Therefore, they give lower performance when the labelled training dataset is small. 

\begin{table*}[htbp]
  \centering
  \caption{Experimental Results on SSW Dataset}
    \begin{tabular}{l|ccc|ccc|ccc|c}
    \toprule
          & \multicolumn{3}{c|}{Sample Average} & \multicolumn{3}{c|}{Weighted Average} & \multicolumn{3}{c|}{Macro Average} &  \\
    \midrule
          & \multicolumn{1}{l}{Accuracy} & \multicolumn{1}{l}{AUC} & \multicolumn{1}{l|}{CKS} & \multicolumn{1}{l}{Precision} & \multicolumn{1}{l}{Recall} & \multicolumn{1}{l|}{F$_1$-score} & \multicolumn{1}{l}{Precision} & \multicolumn{1}{l}{Recall} & \multicolumn{1}{l|}{F$_1$-score} & \multicolumn{1}{l}{Support} \\
    \midrule
    LSTM-LM & \textbf{0.89} & \textbf{0.89} & \textbf{0.77} & \textbf{0.89} & \textbf{0.89} & \textbf{0.89} & \textbf{0.89} & \textbf{0.88} & \textbf{0.89} & 228 \\
    RF    & 0.82  & 0.81  & 0.63  & 0.82  & 0.82  & 0.82  & 0.81  & 0.81  & 0.81  & 228 \\
    DT    & 0.79  & 0.78  & 0.57  & 0.79  & 0.79  & 0.79  & 0.78  & 0.78  & 0.78  & 228 \\
    GNB   & 0.70  & 0.71  & 0.40  & 0.72  & 0.7   & 0.7   & 0.71  & 0.71  & 0.7   & 228 \\
    kNN   & 0.58  & 0.60  & 0.19  & 0.63  & 0.58  & 0.56  & 0.62  & 0.6   & 0.57  & 228 \\
    SVM   & 0.82  & 0.81  & 0.63  & 0.82  & 0.82  & 0.82  & 0.82  & 0.81  & 0.81  & 228 \\
    RC    & 0.77  & 0.77  & 0.53  & 0.77  & 0.77  & 0.77  & 0.77  & 0.77  & 0.77  & 228 \\
    CNN-W & 0.86  & 0.86  & 0.72  & 0.86  & 0.86  & 0.86  & 0.86  & 0.86  & 0.86  & 228 \\
    CNN   & 0.82  & 0.81  & 0.63  & 0.82  & 0.82  & 0.82  & 0.82  & 0.81  & 0.82  & 228 \\
    LSTM-W & 0.85  & 0.84  & 0.70  & 0.85  & 0.85  & 0.85  & 0.86  & 0.84  & 0.85  & 228 \\
    LSTM  & 0.82  & 0.82  & 0.64  & 0.82  & 0.82  & 0.82  & 0.82  & 0.82  & 0.82  & 228 \\
    \bottomrule
    \end{tabular}%
  \label{tab:ssw_results}%
   \vspace{-8mm}
\end{table*}%

\subsection{Experimental Results: The iShield.ai Dataset}
The experimental results comparing our LSTM-LM against ten baseline models on the iShield dataset are given in Table \ref{tab:ishield_results}. 

Accuracy, AUC and CKS are three important measures for understanding the overall significance of a classification model. Table \ref{tab:ishield_results} shows that LSTM-LM gives the best Accuracy, AUC and CKS results. CKS indicates the reliability between the prediction made by a model and the ground truth. All the three transfer learning-based models (i.e. LSTM-LM, LSTM-W and CNN-W) have high CKS value. However, LSTM-LM has better accuracy and AUC than other two (i.e. LSTM-W and CNN-W). 

Overall LSTM-LM gives us the best results, as indicated by best weighted average precision, recall and F$_1$ score. Even though SVM, CNN-W and LSTM-W have the same weighted average precision value as LSTM-W, their recall and F$_1$ score are lower than LSTM-W. 

High precision indicates most of the identified biased sentences are indeed bias. However, if the recall is not high enough, then many bias sentences will left undetected. Therefore, better recall is desirable in bias detection. Yet the excessive false positives can result in a higher cost for investigating many false detection. Therefore, a balance in both recall and precision is needed. A higher $F_1$ score indicates both precision and recall is high. LSTM-LM gives the best F$_1$ score for both weighted average and macro average. Macro Average is used to evaluate the performance of a classifier for minority classes (classes with small number of instances), where weighted average favours majority classes. The high F$_1$-score for both weighted average and macro average indicates that LSTM-LM works reasonably well for both majority and minority classes. Best macro average precision is achieved by SVM, but SVM has very poor recall value and F$_1$-score.



\begin{table*}[htbp]
  \centering
  \caption{Experimental Results on the iShield.ai dataset}
    \begin{tabular}{l|ccc|ccc|ccc|c}
    \toprule
          & \multicolumn{3}{c|}{Sample Average} & \multicolumn{3}{c|}{Weighted Average} & \multicolumn{3}{c|}{Macro Average} &  \\
    \midrule
          & \multicolumn{1}{l}{Accuracy} & \multicolumn{1}{l}{AUC} & \multicolumn{1}{l|}{CKS} & \multicolumn{1}{l}{Precision} & \multicolumn{1}{l}{Recall} & \multicolumn{1}{l|}{F$_1$-score} & \multicolumn{1}{l}{Precision} & \multicolumn{1}{l}{Recall} & \multicolumn{1}{l|}{F$_1$-score} & \multicolumn{1}{l}{Support} \\
    \midrule
    LSTM-LM & \textbf{0.85} & \textbf{0.78} & \textbf{0.73} & \textbf{0.84} & \textbf{0.85} & \textbf{0.84} & 0.69  & \textbf{0.6} & \textbf{0.63} & 5743 \\
    RF    & 0.84  & 0.76  & 0.72  & 0.83  & 0.84  & 0.83  & 0.72  & 0.57  & 0.61  & 5743 \\
    DT    & 0.83  & 0.77  & 0.71  & 0.83  & 0.83  & 0.83  & 0.64  & 0.59  & 0.61  & 5743 \\
    GNB   & 0.62  & 0.74  & 0.45  & 0.79  & 0.62  & 0.67  & 0.44  & 0.58  & 0.43  & 5743 \\
    kNN   & 0.79  & 0.70  & 0.64  & 0.78  & 0.79  & 0.78  & 0.66  & 0.48  & 0.52  & 5743 \\
    SVM   & 0.84  & 0.73  & 0.72  & \textbf{0.84} & 0.84  & 0.83  & \textbf{0.81} & 0.51  & 0.55  & 5743 \\
    RC    & 0.83  & 0.75  & 0.70  & 0.82  & 0.83  & 0.82  & 0.68  & 0.57  & 0.6   & 5743 \\
    CNN-W & 0.84  & 0.76  & \textbf{0.73}  & \textbf{0.84} & 0.84  & 0.83  & 0.73  & 0.57  & 0.59  & 5743 \\
    CNN   & 0.84  & 0.75  & 0.72  & 0.83  & 0.84  & 0.83  & 0.58  & 0.56  & 0.56  & 5743 \\
    LSTM-W & 0.84  & 0.77  & \textbf{0.73}  & \textbf{0.84} & 0.84  & 0.84  & 0.67  & 0.58  & 0.6   & 5743 \\
    LSTM  & 0.84  & 0.76  & 0.72  & 0.84  & 0.84  & 0.83  & 0.69  & 0.58  & 0.6   & 5743 \\
    \bottomrule
    \end{tabular}%
  \label{tab:ishield_results}%
  \vspace{-6mm}
\end{table*}%

\vspace{-2mm}
\section{Deployment and Architecture of Integrated System }
\vspace{-4mm}
The proposed model LSTM-LM is deployed in the flavour of following four different products for the convenience of end users by iShield.ai. 
(1) \textbf{Dost}\footnote{https://ishield.ai/dost}: This bot can be configured to detect bias on enterprise communication platforms such as \emph{Slack} and \emph{Microsoft teams}.
(2) \textbf{Chrome Plugin}\footnote{https://ishield.ai/chrome-plugin}: This plugin can be configured with chrome browser for screening text contents for bias in any web applications. 
(3) \textbf{Content screener}\footnote{https://ishield.ai/screener}: This web application allows Checking for bias in created text contents before publishing them.
(4) \textbf{Application Programming Interface (API)}\footnote{https://ishield.ai/api}: This API can be integrated with enterprise platforms where contents are created and published.

\begin{figure*}[htb!]
\centering
\includegraphics[width=.8\textwidth]{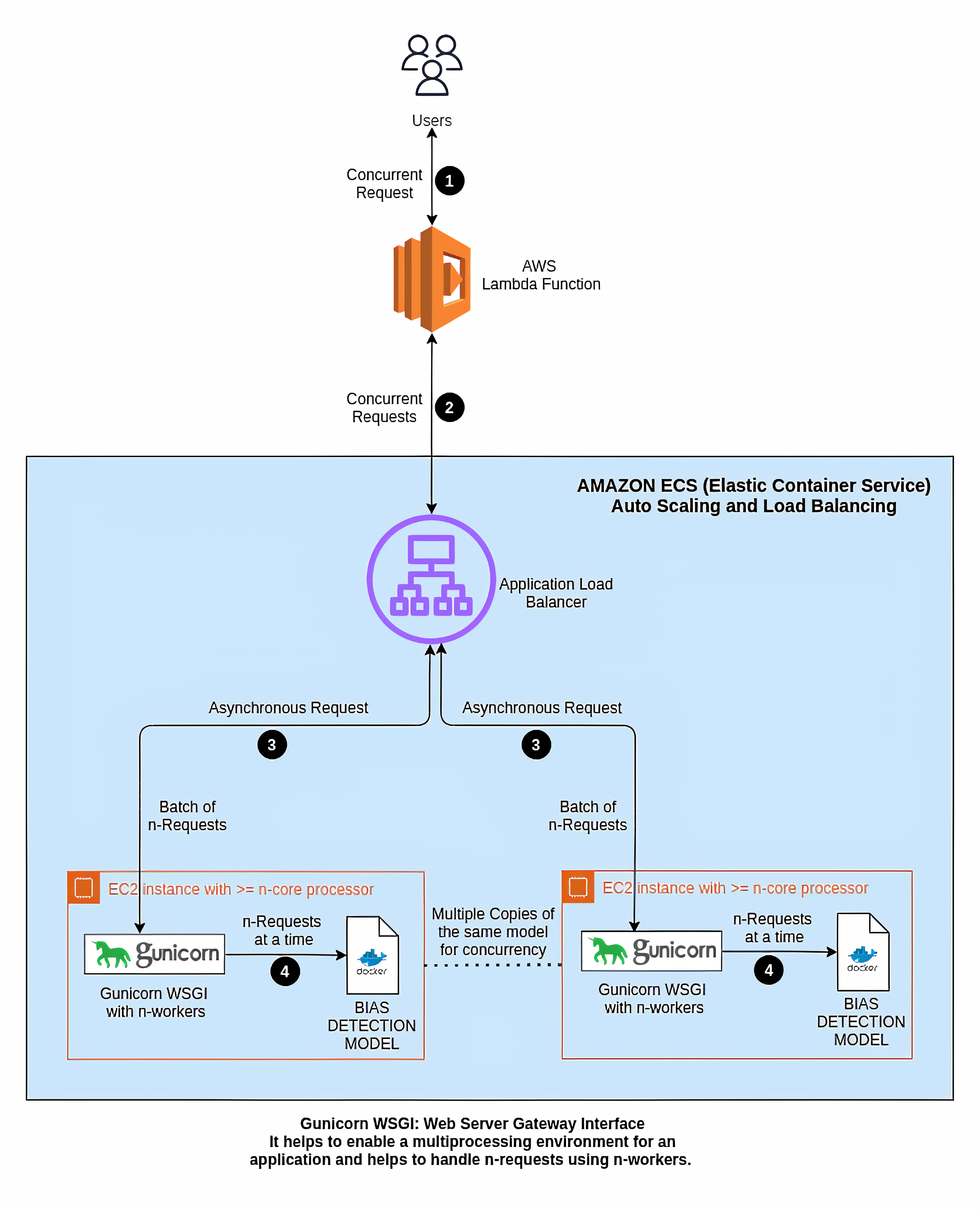}
\caption{Bias Detection System Architecture}
\label{fig:deployment}
\end{figure*}

The backend of the system is deployed on Amazon Web Services (AWS) and is built to scale for industry use. The current architecture can process 8 parallel requests. This architecture can be easily scaled to accommodate increased request volumes. The architecture of the backend system is shown in Fig. \ref{fig:deployment} and can be explained as follows.
(1) Multiple user requests are received concurrently. (2) Requests are received at the AWS Lambda Function. (3) Each model is placed in an n-core EC2 instance. (3.1) Gunicorn Web Server Gateway Interface is implemented with each model to gather and distribute requests for parallel processing. At the time of writing this paper, 8 parallel requests can be processed by each Gunicorn WSGI. This can be scaled up with increased volumes. (3.2) Each model is placed in an independent docker container to isolate it's function from other environment related dependencies. (4) Once a piece of content is identified \emph{biased} by a model, asynchronous requests travel back for confidence based sorting. (5) After index calculation, a response JSON is prepared and sent to the AWS Lamba function. (6) Database operations are performed at the AWS Lamda Function, after which the results travel back to the user.

\vspace{-4mm}
\section{Conclusion}
\vspace{-4mm}
We propose a transfer learning based language model to learn from manually tagged documents for automatically identifying bias in enterprise content in order to create the workplace more inclusive.
The trained model is thoroughly evaluated on independent datasets to ensure a general application, and it is deployed in a real-world application.

%
%
\bibliographystyle{splncs04}
\bibliography{references}

\end{document}